\documentclass[letterpaper]{article} 
\usepackage{aaai2026}  
\usepackage{times}  
\usepackage{helvet}  
\usepackage{courier}  
\usepackage[hyphens]{url}  
\usepackage{graphicx} 
\usepackage{natbib}  
\usepackage{caption} 
\usepackage{booktabs}
\usepackage{amsfonts}
\usepackage{hhline}
\usepackage{algorithm}
\usepackage{algorithmic}
\usepackage{hyperref}
\usepackage{tcolorbox}
\usepackage{newfloat}
\usepackage{listings}
\usepackage{amsmath}
\usepackage{xspace}
\usepackage{marvosym}
\usepackage{gradient-text}
\usepackage{multirow}
\usepackage{colortbl}

\newcommand\NickName{{\gradientRGB{Nav}{255, 192, 0}{226, 102, 102}}${\gradientRGB{A}{255, 192, 0}{226, 102, 102}}^{\gradientRGB{3}{255, 192, 0}{226, 102, 102}}$\xspace}
\definecolor{myred}{RGB}{226, 102, 102}
\newcommand\website{\gradientRGB{https://NavigationA3.github.io/}{255, 192, 0}{226, 102, 102}}

\DeclareCaptionStyle{ruled}{labelfont=normalfont,labelsep=colon,strut=off} 
\floatstyle{ruled}
\newfloat{listing}{tb}{lst}{}
\floatname{listing}{Listing}
\definecolor{lightorangered1}{RGB}{255,140,105}
\title{\NickName: Understanding \textcolor{lightorangered1}{Any} Instruction, Navigating \textcolor{lightorangered1}{Any}where, Finding \textcolor{lightorangered1}{Any}thing}

\author{
Lingfeng Zhang$^{1,2,*}$ ~~~~~~~ Xiaoshuai Hao$^{2,*,\dag}$ ~~~~~~~ Yingbo Tang$^{2,3}$ ~~~~~~~ Haoxiang Fu$^{6}$ \\
{\bf Xinyu Zheng$^{4}$} ~~~~~ {\bf Pengwei Wang$^{2}$} ~~~~~ {\bf Zhongyuan Wang$^{2}$} \\
{\bf Wenbo Ding$^{1,}$\textsuperscript{\Letter}} ~~~~~ {\bf Shanghang Zhang$^{2,5,}$\textsuperscript{\Letter}}
}

\affiliations{{\textsuperscript{\rm 1}Tsinghua University} \\{$^2$ Beijing Academy of Artificial Intelligence (BAAI)} \\ {$^3$Institute of Automation, Chinese Academy of Sciences} \\ {$^4$ National Maglev Transportation Engineering Research and Development Center, Tongji University} \\ {$^5$ State Key Laboratory of Multimedia Information Processing, School of Computer Science, Peking University} \\ {$^6$ National University of Singapore} \\
\small $^*$ Co-first Authors, $^\dag$ Project Leader, \textsuperscript{\Letter} Corresponding Authors \\
\texttt{lfzhang715@gmail.com, xshao@baai.ac.cn, ding.wenbo@sz.tsinghua.edu.cn, shanghang@pku.edu.cn}
}

\usepackage{bibentry}

\begin{document}

\thispagestyle{plain}

\twocolumn[{
\renewcommand\twocolumn[1][]{#1}
\vspace{-20pt}
\maketitle

\vspace{-3pt}
\begin{center}
    \captionsetup{type=figure}
    \includegraphics[width=0.94\textwidth]{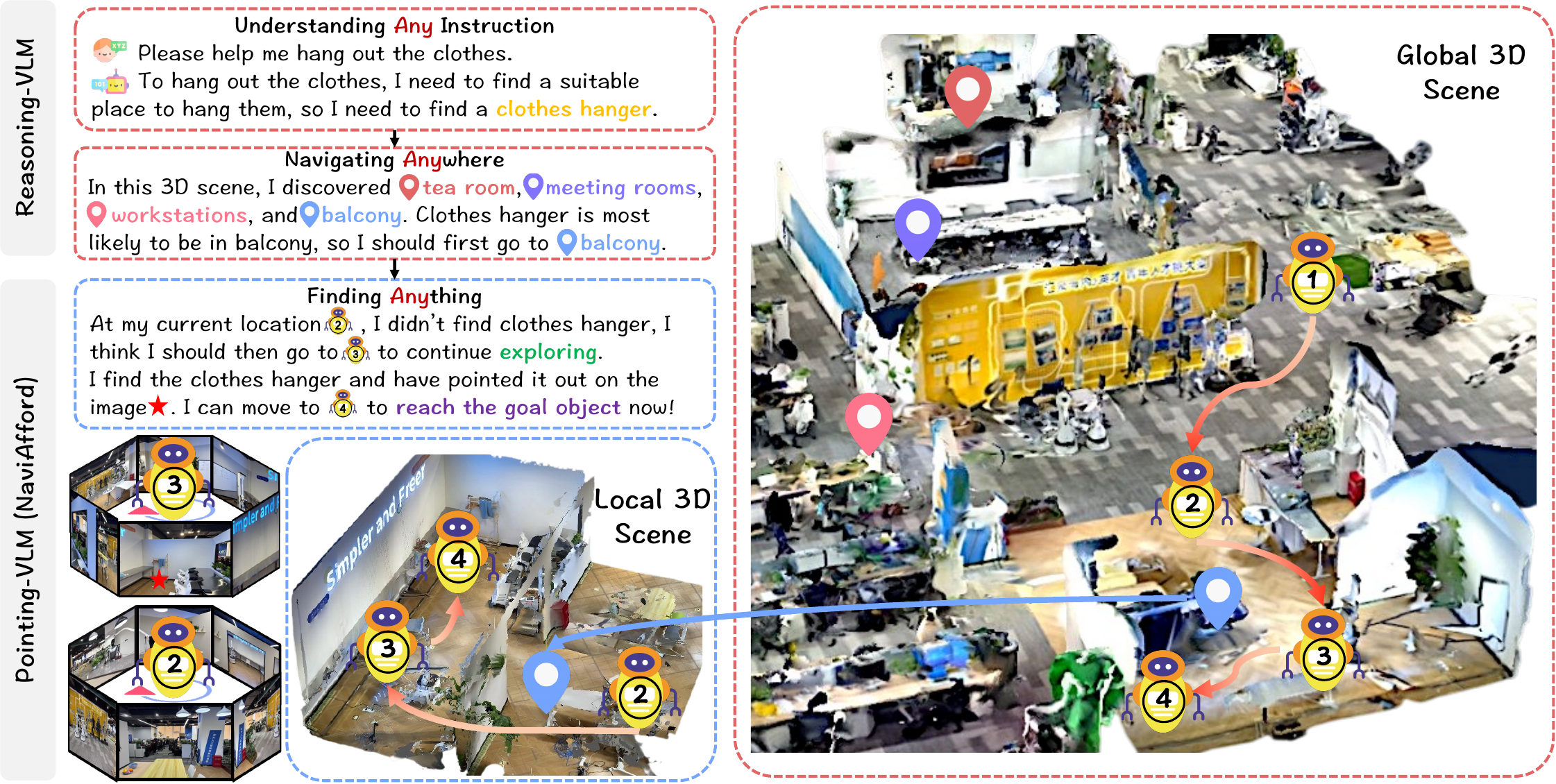}
\captionof{figure}{\textbf{Execution Process of \textit{\NickName}.} The global policy employs \textbf{\textit{Reasoning-VLM}} to interpret high-level instructions (\textit{e.g.}, ``hang out the clothes'' $\rightarrow$ clothes hanger) and identify the target location (balcony) using 3D scene understanding. The local policy uses \textbf{\textit{Pointing-VLM}} to navigate waypoints and perform precise object localization with our \textit{\textbf{NaviAfford}} model, which leverages spatial affordance understanding to accurately locate the target object (clothes hanger).}
    \label{fig1}
\end{center}

}]

\pagestyle{empty}

\begin{abstract}

Embodied navigation is a fundamental capability of embodied intelligence, enabling robots to move and interact within physical environments. However, existing navigation tasks primarily focus on predefined object navigation or instruction following, which significantly differs from human needs in real-world scenarios involving complex, open-ended scenes.
To bridge this gap, we introduce a challenging \emph{long-horizon navigation task} that requires understanding high-level human instructions and performing spatial-aware object navigation in real-world environments. Existing embodied navigation methods struggle with such tasks due to their limitations in comprehending high-level human instructions and localizing objects with an open vocabulary.
In this paper, we propose \textit{\textbf{\NickName}}, a hierarchical framework divided into two stages: global and local policies. In the global policy, we leverage the reasoning capabilities of \textit{\textbf{Reasoning-VLM}} to parse high-level human instructions and integrate them with global 3D scene views. This allows us to reason and navigate to regions most likely to contain the goal object. In the local policy, we have collected a dataset of 1.0 million samples of spatial-aware object affordances to train the \textit{\textbf{NaviAfford}} model (\textit{\textbf{Pointing-VLM}}), which provides robust open-vocabulary object localization and spatial awareness for precise goal identification and navigation in complex environments.
Extensive experiments demonstrate that \textit{\textbf{\NickName}} achieves SOTA results in navigation performance and can successfully complete \emph{long-horizon navigation tasks} across different robot embodiments in real-world settings, paving the way for universal embodied navigation. The dataset and code will be made available. Project website: {\textit{\website}}.
\end{abstract}

\section{Introduction}
\label{sec1}
Embodied navigation~\cite{zheng2024towards, morad2021embodied} is a foundational capability of embodied intelligence, essential for robots to perform complex tasks in physical environments. This capability enables autonomous agents to navigate and interact within real-world spaces, forming the basis for more sophisticated embodied behaviors such as manipulation, exploration, and human-robot collaboration~\cite{hao2025mapfusion, tang2025affordgrasp, zhang2024multi, tang2025affordgrasp, hao2024mapdistill, hao2025msc, hao2024mbfusion, li2024foundation, hao2025safemap, zhang2025video, zhang2025vtla, hao2025tla,zhao2025training, zhang2025humanoidpano}. Despite significant advancements in this field, existing research primarily focuses on relatively low-level tasks, such as instruction following and basic object navigation, which do not fully capture the nuances of human needs in dynamic environments.
Current embodied navigation methods can be broadly categorized into two main approaches: vision and language navigation (VLN)~\cite{hong2021vln, zheng2025railway, chen2024mapgpt} and object navigation (ObjectNav)~\cite{ cai2024bridging, qi2025vln, gao2025octonav, gong2025stairway}. VLN tasks require agents to follow detailed, step-by-step instructions, such as ``turn left, go out the door, and then go straight." While these tasks necessitate precise spatial understanding, they often rely on overly specific directives that are seldom provided by humans in natural settings. Conversely, ObjectNav tasks aim to locate predefined object categories (\textit{e.g.}, ``find any chair in the scene") and succeed upon encountering any instance of the target object, regardless of the spatial context or specific requirements.

However, real-world human instructions frequently involve high-level intentions that demand complex reasoning and spatial perception. For instance, requests like ``I want a cup of coffee" or ``I want to eat the fruit on the left side of the tea room" necessitate not only an understanding of the underlying targets but also reasoning about the spatial relationships among objects. This highlights a fundamental gap between current navigation tasks and real-world needs, which significantly hampers the development of embodied agents capable of advanced human-computer interaction.

To address the challenges of long-horizon navigation, we propose \textit{\textbf{\NickName}}, a novel hierarchical framework that decomposes this complex problem into two stages: \emph{global policy and local policy.}
As shown in Fig.~\ref{fig1}, the global policy leverages the powerful reasoning capabilities of the vision language model (VLM)~\cite{ji2025robobrain, o2024open, tan2025reason,zhai2024fine}, termed \textit{\textbf{Reasoning-VLM}}, to parse high-level human instructions. Reasoning-VLM identifies key objects to locate based on these instructions and determines the most probable space for the goal object using an annotated global 3D scene. For example, when given the instruction ``I want a cup of coffee,'' the global policy infers that the coffee machine is likely located in the pantry, guiding the agent to this high-probability area.
Upon completion of the global policy, the local policy takes over, focusing on exploration and precise object localization within the identified target area. The VLM, referred to as \textit{\textbf{Pointing-VLM}}, selects waypoints from the local 3D scene for exploration. At each waypoint, we perform panoramic perception and utilize our specially trained \textit{\textbf{NaviAfford}} model (an implementation of Pointing-VLM) for accurate target object identification. If the target object is detected, we transform its location from the agent's perspective to the robot's coordinate system, enabling navigation to the final goal.
The \textit{\textbf{NaviAfford}} model is trained on a spatial object affordance dataset comprising 1.0 million sample pairs, facilitating spatial-aware object and affordance localization. This allows the model to understand complex spatial relationships, such as ``cup by the window'' or ``empty space on the left side of the table.''
Extensive experimental evaluations demonstrate that \textit{\textbf{\NickName}} achieves state-of-the-art performance in long-horizon navigation tasks across large-scale real-world environments. Additionally, our system exhibits excellent cross-embodiment capabilities, making it adaptable to various robot instances and highlighting its potential for practical applications. 


Our contributions are summarized as follows:
\begin{itemize}
    \item We introduce a challenging and realistic long-horizon navigation task that requires agents to comprehend high-level human instructions and locate open vocabulary objects with complex spatial relationships in intricate indoor environments.
    \item We propose \textit{\textbf{\NickName}}, a novel hierarchical framework leveraging both global and local policies. This framework enables the understanding of diverse high-level instructions, navigation across various areas, and the ability to find any object.
    \item We have collected a dataset of 1.0 million samples of spatial-aware object affordances to train the \textit{\textbf{NaviAfford}} model, enabling it to effectively understand complex spatial relationships and perform accurate object pointing.
    \item Extensive experiments demonstrate that our approach achieves state-of-the-art navigation performance compared to existing methods, paving the way for the development of general embodied navigation systems in real-world scenarios.
\end{itemize}

\vspace{-10pt}
\section{Related Work}
\begin{figure*}[!ht]
\centering
\includegraphics[width=0.93\textwidth]{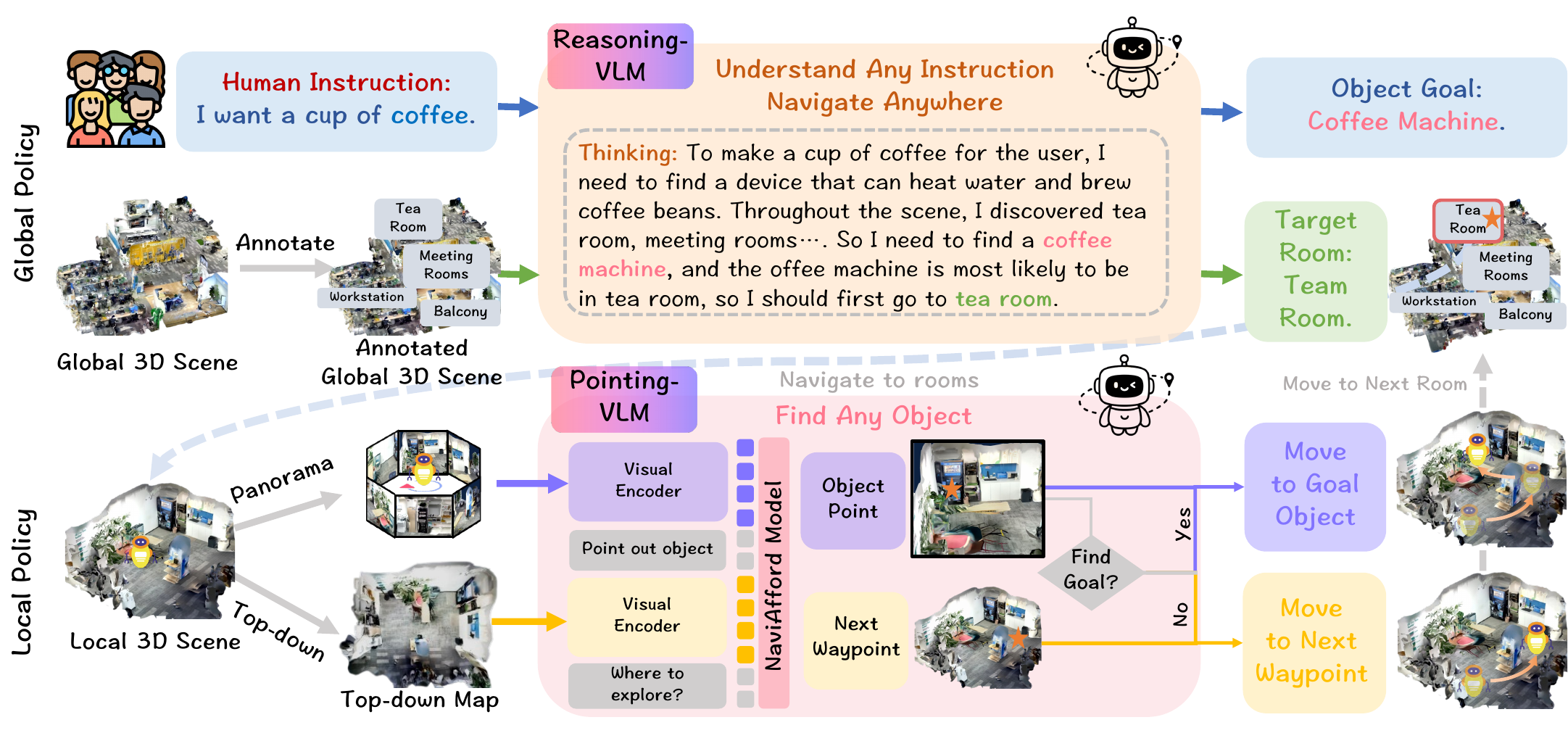} 
\caption{\textbf{Overview of the \textit{\NickName} Framework.} Our hierarchical approach consists of two stages: the global policy uses Reasoning-VLM to interpret high-level human instructions and marks the probable area in the 3D scene. Upon reaching the target area, the local policy employs Pointing-VLM to search for the goal object at each waypoint. If not found, it predicts the next waypoint; if detected, it marks the object on the egocentric image and navigates to the final destination.}
\label{fig2}
\vspace{-1em}
\end{figure*}

\textbf{Embodied Navigation}
Embodied navigation research focuses on two main paradigms: visual-language navigation (VLN) and object-target navigation (ObjectNav)~\cite{chattopadhyay2021robustnav, truong2021bi}. In VLN, systems like NavGPT~\cite{zhou2024navgpt} utilize GPT-4o~\cite{hurst2024gpt} for autonomous action generation, while DiscussNav~\cite{long2024discuss} reduces human involvement. InstructNav~\cite{long2024instructnav} breaks navigation into subtasks, and Nav-CoT~\cite{lin2025navcot} uses chain-of-thought reasoning for simulations. MapNav~\cite{zhang2025mapnav} optimizes memory with spatial representations, and NaVid~\cite{zhang2024navid} maintains temporal context. For ObjectNav, methods like PirlNav~\cite{ramrakhya2023pirlnav} and XGX~\cite{wasserman2024exploitation} imitate human demonstrations, while others such as L3MVN~\cite{yu2023l3mvn} and Uni-NaVid~\cite{zhang2024uni} build semantic maps or leverage VLMs for enhanced performance. However, these approaches primarily focus on detailed instructions and lack the ability to understand high-level human commands or perform spatial-aware localization of open-vocabulary objects, limiting their effectiveness in long-horizon navigation tasks.

\textbf{Spatial Reasoning with VLMs}
Spatial reasoning is vital for robots to interact with the physical world~\cite{beyer2024paligemma, wang2023visionllm, luo2023cheap, liu2024vision, liu2023qilin, doveh2024towards}. Researchers have developed various methods to enhance the spatial understanding of VLMs by extracting spatial information from images. For example, SpatialVLM~\cite{chen2024spatialvlm} converts images into object-centered point clouds, while SpatialRGPT~\cite{cheng2024spatialrgpt} enhances region-level reasoning through spatial scene graphs. RoboPoint~\cite{yuan2024robopoint} introduces a synthetic dataset for precise action predictions, and studies like SpatialBot~\cite{cai2024spatialbot} use RGB-D data for comprehensive spatial understanding. Recent advancements, such as SpatialCoT~\cite{liu2025spatialcot} and VILASR~\cite{wu2025reinforcing}, focus on optimizing reasoning processes. However, these methods still struggle with open-vocabulary spatial-aware object pointing and long-horizon navigation integration, which are crucial for practical applications.

\section{Methodology}
As shown in Fig.~\ref{fig2}, our \NickName framework employs a hierarchical global-to-local approach that integrates semantic reasoning with precise spatial localization to tackle long-view navigation tasks. The global policy utilizes \emph{Reasoning-VLM} to interpret high-level human instructions (\textit{e.g.}, ``I want a cup of coffee''), inferring the target object (coffee machine) and identifying the likely room (\textit{e.g.}, tea room, kitchen area). Upon reaching this room, the local policy uses our \textbf{\textit{NaviAfford}} model (\emph{Pointing-VLM}) to analyze panoramic RGB observations and the local map at each waypoint. The model determines if the target object is present; if found, it points to its location for navigation, and if not, it predicts the next best waypoint or consults Reasoning-VLM to continue exploration until the target is located.

\begin{figure*}[!t]
\centering
\includegraphics[width=0.94\textwidth]{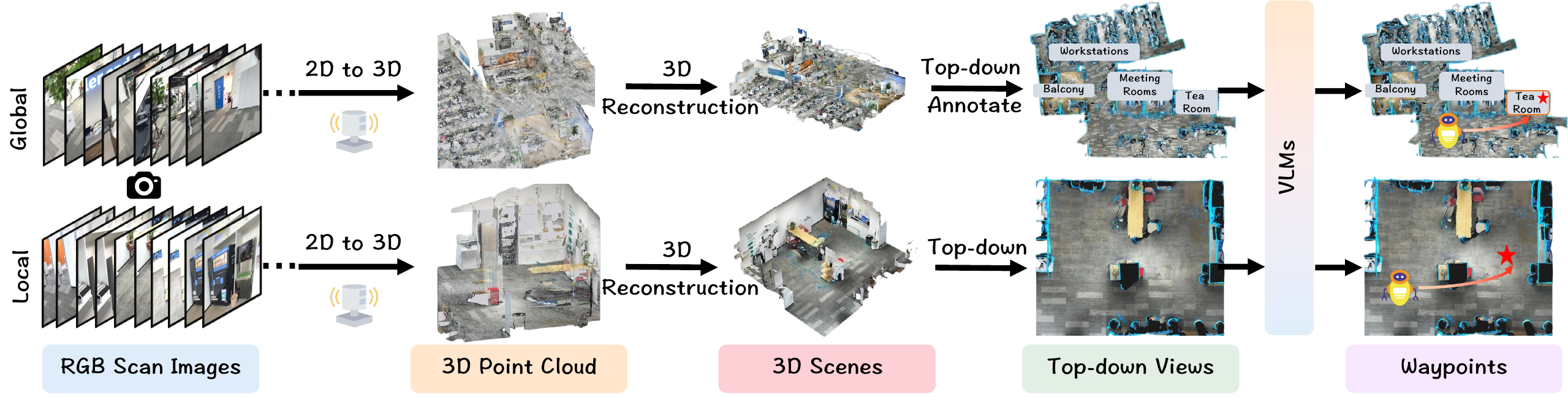} 
\caption{\textbf{Construction Process of 3D Scenes.} We reconstruct 3D scenes from RGB scan images using 2D-to-3D reconstruction techniques. These scenes are then transformed into annotated top-down views, which are subsequently processed by Vision-Language Models (VLMs) for navigation planning. This approach enhances the accuracy and efficiency of navigation tasks.}
\label{fig3}
\vspace{-1em}
\end{figure*}

\subsection{Preliminaries}
\textbf{Problem Definition} We formulate the long-horizon embodied navigation task as follows: given a high-level human instruction $I$ (\textit{e.g.}, ``I want coffee'' or ``Help me hang the clothes on the balcony''), the embodied agent must navigate within a large indoor environment $E$ to locate and reach a specific target object $O$ that fulfills the semantic and spatial requirements implied by the instruction. The agent starts at an arbitrary position $p_0$ and has access to egocentric RGB-D observations $o_t$ and a global 3D scene representation $S$. Unlike traditional ObjectNav tasks, which terminate upon finding any instance of a predefined object category, our task necessitates multi-step reasoning to infer specific target objects from high-level instructions (\textit{e.g.}, inferring ``coffee machine'' from ``I want coffee''), identifying the most likely spatial locations (\textit{e.g.}, kitchen or tea room), and navigating to the precise object instance that meets the contextual requirements. Success is defined as the agent reaching within 1 meter of the target object $O$ while maintaining line of sight, showcasing accurate semantic understanding and precise spatial navigation in complex real-world environments.

\textbf{3D Scenes Construction} To enable effective navigation in real-world environments, we construct a hierarchical 3D scene representation using a straightforward reconstruction pipeline, as illustrated in Fig.~\ref{fig3}. Our process begins with a sequence of RGB images captured from multiple viewpoints, which are processed through a 2D-to-3D reconstruction pipeline. Using a mobile device equipped with a LiDAR sensor, we generate a dense point cloud represented by:
\begin{equation}
   P = \{p_i | p_i \in \mathbb{R}^3\}_{i=1}^N,
\end{equation}
where each point $p_i$ represents a 3D coordinate in the scene. The reconstruction employs a feature point matching algorithm to establish correspondences between consecutive frames, followed by mesh reconstruction to generate coherent 3D geometry. A 3D scanner application is used to streamline this process and ensure high-quality results.

The reconstructed 3D scene is converted into a top-down view for global and local policies. For the global policy, we use MapNav's~\cite{zhang2025mapnav} annotation method to provide room- and region-level semantic annotations, such as ``tea room'', ``conference room'', ``balcony'', and ``workstation''. This allows the VLM to effectively understand spatial semantics and reason about object locations. The annotated global scene is represented as:
\begin{equation}
S_{\text{global}} = \{R_j, A_j\}_{j=1}^M,
\end{equation}
where $R_j$ represents the geometric region and $A_j$ the corresponding semantic annotation. For the local strategy, we use the top-down map $M_{\text{local}}$ directly, without annotations.

\subsection{Global Policy}
The global policy utilizes the advanced reasoning capabilities of the vision language model (Reasoning-VLM) to bridge the semantic gap between high-level human instructions and navigation goals. As shown in Fig.~\ref{fig2}, given human instructions $I$ and an annotated global 3D scene $S_{\text{global}}$, we treat the global reasoning task as a multimodal problem where Reasoning-VLM performs both semantic object reasoning and spatial location prediction.

\emph{To support systematic reasoning, we designed a structured prompt template to effectively guide Reasoning-VLM:}
\begin{tcolorbox}
\texttt{``You need to complete the human instruction: $I$. Now given this top-down scene view $S_{global}$ and several optional regions, please think about what object you should find to complete the instruction and where you should look for this object. Please show your thinking process and give your answer at the end.''}
\end{tcolorbox}

The Reasoning-VLM processes textual instructions and the visual representation of the annotated global scene to enable hierarchical reasoning. It first infers the target object $O^*$ needed to fulfill the instruction through semantic decomposition: $O^* = f_{\text{semantic}}(I)$. The model then analyzes spatial semantic relations to identify the target region $R^*$, where the object is most likely located, defined by $R^* = \arg\max_{R_j \in S_{\text{global}}} P(O^* | R_j, A_j)$, with $P(O^* | R_j, A_j)$ representing the conditional probability of finding $O^*$ with annotation $A_j$ in region $R_j$.

After identifying the target region $R^*$, we randomly sample a waypoint $w \in R^*$ within its local boundary and use Pointing-VLM to guide the agent. This strategy promotes robust exploration while effectively narrowing the search space to relevant subregions where the target object is likely located, enhancing the efficiency of the search process.

\begin{figure*}[!t]
\centering
\includegraphics[width=0.97\textwidth]{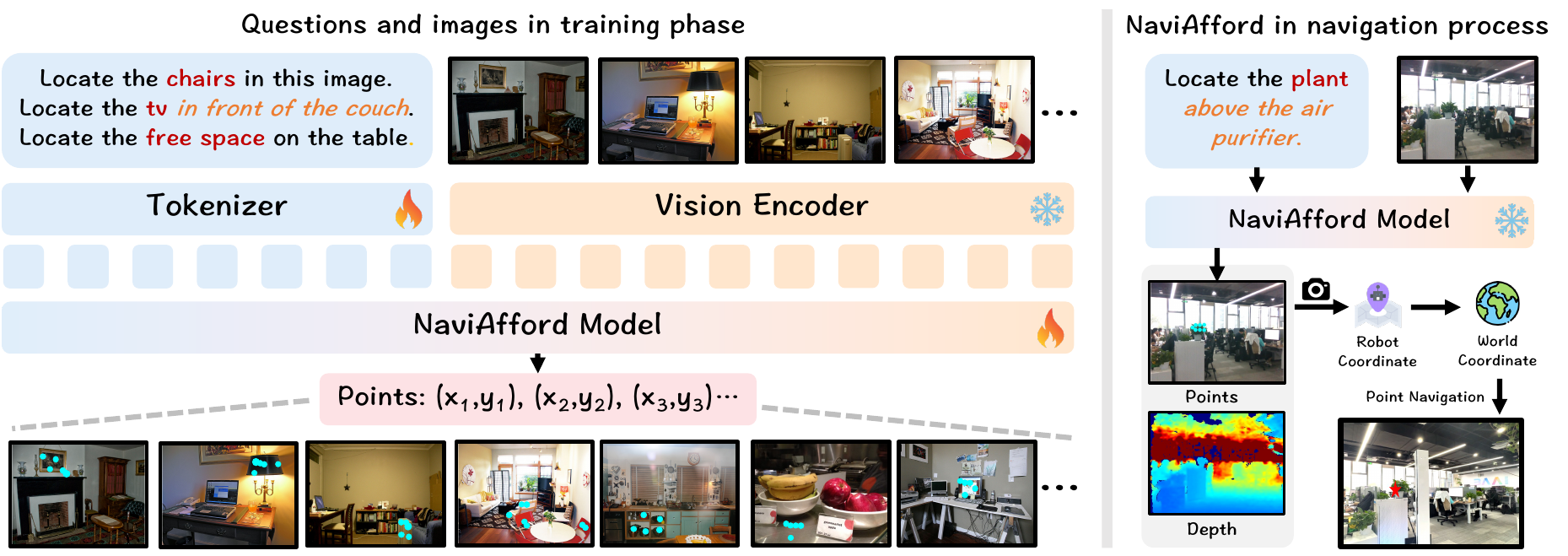} 
\vspace{-5pt}

\caption{\textbf{\textit{NaviAfford} Model Training and Deployment Process.} The \textit{\textbf{NaviAfford}} model learns object and spatial affordances from various indoor scenes to output precise point coordinates. During navigation, it performs real-time object localization and generates target points, which the local policy converts into robot coordinates for effective navigation to goal objects.}
\label{fig4}
\vspace{-1.5em}
\end{figure*}

\subsection{Local Policy}
\textbf{NaviAfford Model}  
To achieve precise spatial object localization, we developed the NaviAfford model (Pointing-VLM), as shown in Fig.~\ref{fig4}. During training, we curated a dataset of approximately 50K images and 1.0M question-answer pairs from the LVIS and Where2Place datasets. We converted instance segmentation masks into an object detection format with bounding box coordinates $(x_1, y_1, x_2, y_2)$ and sampled 5-8 representative points within each box to enhance spatial granularity and improve localization accuracy, supporting Reasoning-VLM's capabilities.

Our dataset construction systematically generates two types of affordance annotations for comprehensive spatial understanding. For \textbf{\textit{object affordance}}, we compute directional relations (up, down, left, right, front, back) to identify target objects in specific contexts. For example, given the query "find the TV in front of the sofa," we determine the target object and its spatial relationship to reference objects. For \textit{\textbf{spatial affordance}}, we identify free spaces that satisfy these constraints, enabling the model to understand available areas for navigation and placement. This dual-affordance approach creates training samples that capture the complex spatial relations necessary for real-world navigation.

The NaviAfford model architecture follows a vision-language framework, processing the input question \( Q \) and RGB image \( V \) through separate tokenizer and visual encoder paths. The architecture is expressed as:
\begin{equation}
    \text{\textit{NaviAfford}}(Q, V) = f_{\text{LLM}}(f_{\text{text}}(Q), f_{\text{proj}}(f_{\text{vision}}(V))),
\end{equation}
where \( f_{\text{text}} \) processes the text query, \( f_{\text{vision}} \) encodes the visual input, and \( f_{\text{proj}} \) maps visual features to the LLM embedding space. The function \( f_{\text{LLM}} \) generates text point coordinates. The training objective uses supervised fine-tuning (SFT) with the loss function:
\begin{equation}
    \mathcal{L} = -\sum_{i=1}^{N} \log P(t_i | t_{<i}, Q, V),
\end{equation}
where \( t_i \) represents the \( i \)-th token containing point coordinates in the target text sequence. In the local policy of \textit{\textbf{NaviAfford}}, we input the self-centered RGB view and the target object query based on spatial relationships, deploying the model in real-world environments with zero samples. The model outputs accurate point coordinates, with specific usage detailed in the local policy.

\begin{table*}[!ht]
\renewcommand{\arraystretch}{1.5}
\centering
\caption{\textbf{Navigation Performance Comparison with SOTA methods.} $^*$ denotes that we modify the method to allow it to complete our task. Our \textbf{\textit{\NickName}} outperforms all the SOTA methods on the navigation performance.}
\label{tab:navigation_results}
\resizebox{0.93\textwidth}{!}{%
\begin{tabular}{l|l|cc|cc|cc|cc|cc|c}
\hline
\multirow{2}{*}{\textbf{Category}} & \multirow{2}{*}{\textbf{Methods}} & \multicolumn{2}{c|}{\textbf{Meeting Room A}} & \multicolumn{2}{c|}{\textbf{Meeting Room B}} & \multicolumn{2}{c|}{\textbf{Tea Room}} & \multicolumn{2}{c|}{\textbf{Workstation}} & \multicolumn{2}{c|}{\textbf{Balcony}} & \multirow{2}{*}{\textbf{Avg. SR$\uparrow$}} \\
\hhline{~~----------}
& & \cellcolor{gray!6}\textbf{NE$\downarrow$} & \cellcolor{gray!6}\textbf{SR$\uparrow$} & \cellcolor{gray!6}\textbf{NE$\downarrow$} & \cellcolor{gray!6}\textbf{SR$\uparrow$} & \cellcolor{gray!6}\textbf{NE$\downarrow$} & \cellcolor{gray!6}\textbf{SR$\uparrow$} & \cellcolor{gray!6}\textbf{NE$\downarrow$} & \cellcolor{gray!6}\textbf{SR$\uparrow$} & \cellcolor{gray!6}\textbf{NE$\downarrow$} & \cellcolor{gray!6}\textbf{SR$\uparrow$} & \\
\hline
\multirow{3}{*}{\parbox{0.7cm}{Closed-source}} & \cellcolor{red!6}GPT-4o~\cite{hurst2024gpt} & \cellcolor{red!6}12.45 & \cellcolor{red!6}2.0\% & \cellcolor{red!6}13.78 & \cellcolor{red!6}0.0\% & \cellcolor{red!6}14.12 & \cellcolor{red!6}2.0\% & \cellcolor{red!6}11.89 & \cellcolor{red!6}4.0\% & \cellcolor{red!6}10.45 & \cellcolor{red!6}2.0\% & \cellcolor{red!6}2.0\% \\
& \cellcolor{red!6}Claude-3.5-Sonnet~\cite{anthropic2024claude} & \cellcolor{red!6}11.18 & \cellcolor{red!6}6.0\% & \cellcolor{red!6}12.56 & \cellcolor{red!6}0.0\% & \cellcolor{red!6}13.94 & \cellcolor{red!6}2.0\% & \cellcolor{red!6}10.67 & \cellcolor{red!6}4.0\% & \cellcolor{red!6}11.31 & \cellcolor{red!6}2.0\% & \cellcolor{red!6}2.8\% \\
& \cellcolor{red!6}Qwen-VL-Max~\cite{bai2025qwen2} & \cellcolor{red!6}13.67 & \cellcolor{red!6}0.0\% & \cellcolor{red!6}15.01 & \cellcolor{red!6}0.0\% & \cellcolor{red!6}16.45 & \cellcolor{red!6}0.0\% & \cellcolor{red!6}14.12 & \cellcolor{red!6}2.0\% & \cellcolor{red!6}12.89 & \cellcolor{red!6}0.0\% & \cellcolor{red!6}0.4\% \\
\hline
\multirow{3}{*}{\parbox{0.7cm}{Open-source}}& \cellcolor{yellow!6}Janus-Pro-7B~\cite{chen2025janus} & \cellcolor{yellow!6}16.42 & \cellcolor{yellow!6}0.0\% & \cellcolor{yellow!6}17.89 & \cellcolor{yellow!6}0.0\% & \cellcolor{yellow!6}18.23 & \cellcolor{yellow!6}0.0\% & \cellcolor{yellow!6}15.98 & \cellcolor{yellow!6}0.0\% & \cellcolor{yellow!6}16.54 & \cellcolor{yellow!6}0.0\% & \cellcolor{yellow!6}0.0\% \\
& \cellcolor{yellow!6}Qwen2.5-VL-7B~\cite{bai2025qwen2} & \cellcolor{yellow!6}18.98 & \cellcolor{yellow!6}0.0\% & \cellcolor{yellow!6}19.34 & \cellcolor{yellow!6}0.0\% & \cellcolor{yellow!6}20.78 & \cellcolor{yellow!6}0.0\% & \cellcolor{yellow!6}17.45 & \cellcolor{yellow!6}0.0\% & \cellcolor{yellow!6}18.12 & \cellcolor{yellow!6}0.0\% & \cellcolor{yellow!6}0.0\% \\
& \cellcolor{yellow!6}LLaVA-Next-7B~\cite{liu2024llavanext} & \cellcolor{yellow!6}17.98 & \cellcolor{yellow!6}0.0\% & \cellcolor{yellow!6}18.34 & \cellcolor{yellow!6}0.0\% & \cellcolor{yellow!6}19.78 & \cellcolor{yellow!6}0.0\% & \cellcolor{yellow!6}16.45 & \cellcolor{yellow!6}0.0\% & \cellcolor{yellow!6}17.12 & \cellcolor{yellow!6}0.0\% & \cellcolor{yellow!6}0.0\% \\
\hline
\multirow{4}{*}{\parbox{1.5cm}{Navigation-specific}} & \cellcolor{green!6}NaVid$^*$~\cite{zhang2024navid} & \cellcolor{green!6}8.14 & \cellcolor{green!6}18.0\% & \cellcolor{green!6}9.31 & \cellcolor{green!6}12.0\% & \cellcolor{green!6}10.52 & \cellcolor{green!6}10.0\% & \cellcolor{green!6}7.89 & \cellcolor{green!6}16.0\% & \cellcolor{green!6}8.76 & \cellcolor{green!6}18.0\% & \cellcolor{green!6}14.8\% \\
& \cellcolor{green!6}NaVILA$^*$~\cite{cheng2024navila} & \cellcolor{green!6}7.93 & \cellcolor{green!6}20.0\% & \cellcolor{green!6}8.67 & \cellcolor{green!6}16.0\% & \cellcolor{green!6}9.84 & \cellcolor{green!6}12.0\% & \cellcolor{green!6}7.45 & \cellcolor{green!6}18.0\% & \cellcolor{green!6}8.22 & \cellcolor{green!6}16.0\% & \cellcolor{green!6}16.4\% \\
& \cellcolor{green!6}MapNav$^*$~\cite{zhang2025mapnav} & \cellcolor{green!6}7.21 & \cellcolor{green!6}26.0\% & \cellcolor{green!6}7.94 & \cellcolor{green!6}24.0\% & \cellcolor{green!6}9.12 & \cellcolor{green!6}26.0\% & \cellcolor{green!6}6.78 & \cellcolor{green!6}28.0\% & \cellcolor{green!6}7.45 & \cellcolor{green!6}22.0\% & \cellcolor{green!6}25.2\% \\
\hhline{~------------}
& \cellcolor{blue!6}\textbf{\NickName (Ours)} & \cellcolor{blue!6}\textbf{1.23} & \cellcolor{blue!6}\textbf{72.0\%} & \cellcolor{blue!6}\textbf{1.45} & \cellcolor{blue!6}\textbf{64.0\%} & \cellcolor{blue!6}\textbf{1.89} & \cellcolor{blue!6}\textbf{60.0\%} & \cellcolor{blue!6}\textbf{1.56} & \cellcolor{blue!6}\textbf{76.0\%} & \cellcolor{blue!6}\textbf{1.34} & \cellcolor{blue!6}\textbf{60.0\%} & \cellcolor{blue!6}\textbf{66.4\%} \\
\hline
\end{tabular}%
}
\vspace{-1.5em}
\end{table*}

\textbf{Navigation Process}  
In the local policy, our system employs a fine-grained object localization and navigation strategy based on systematic waypoint exploration. As shown in Fig.~\ref{fig2}, the agent captures panoramic RGB views at each waypoint via rotational scanning. The NaviAfford model processes these views to detect and accurately localize the target object. Upon detection, the model outputs multiple point coordinates, and we select the center point by averaging for robust localization. 

To convert pixel coordinates to robot coordinates, we use the camera intrinsic function:
\begin{equation}
    \begin{bmatrix} X \\ Y \\ Z \end{bmatrix} = \begin{bmatrix} \frac{(u - c_x) \cdot d}{f_x} \\ \frac{(v - c_y) \cdot d}{f_y} \\ d \end{bmatrix},
\end{equation}
where \(f_x\) and \(f_y\) are the focal lengths, \(c_x\) and \(c_y\) are the principal points, and \(d\) is the depth at pixel \((u, v)\). This ensures effective navigation to the target object.

Next, we transform camera coordinates to robot coordinates using rotation and translation:
\begin{equation}
    \begin{bmatrix} x_{world} \\ y_{world} \end{bmatrix} = \begin{bmatrix} \cos\theta_r & -\sin\theta_r \\ \sin\theta_r & \cos\theta_r \end{bmatrix} \begin{bmatrix} x_{robot} \\ y_{robot} \end{bmatrix} + \begin{bmatrix} x_r \\ y_r \end{bmatrix},
\end{equation}
where \((x_r, y_r, \theta_r)\) is the robot's world pose, and \((x_{robot}, y_{robot})\) are derived from camera coordinates: \(x_{robot} = Z_{cam}, y_{robot} = -X_{cam}\).

If the goal object is not detected, the system follows a two-stage decision process. First, Reasoning-VLM analyzes the local 3D scene and historical exploration data to decide whether to continue exploring the current area or transition to a new one. If it opts to continue, the NaviAfford model identifies the next best exploration point. Otherwise, it selects the most promising room or space to explore based on previous searches, facilitating efficient transitions.

\section{Experiments}
\subsection{Experimental Details}
\textbf{Evaluation Benchmark}  
To assess long-horizon navigation performance, we established a benchmark with five distinct scenes: Meeting Room A, Meeting Room B, Tea Room, Workstation, and Balcony. Each scene includes 10 navigation tasks, totaling 50 tasks. For each method, we conducted 10 rollouts per task to minimize randomness. Human experts defined high-level instructions and associated semantic objects, ensuring the uniqueness of the goal object in each scene. Each task was tested five times under varying starting conditions for reliability. During execution, the agent freely interacts with the environment, utilizing egocentric RGB-D perception, waypoint selection, and action control. For evaluating different PointVLM models, we selected 1,000 images not present in the training set.

\textbf{Evaluation Metrics}  
We employed two standard metrics in embodied navigation: Navigation Error (NE) and Success Rate (SR). NE measures the Euclidean distance (in meters) between the agent’s final position and the target, with lower values indicating better performance. SR reflects the percentage of navigation events where the agent successfully reaches the target, defined as being within 1 meter. We compute SR across 10 tasks, each tested 5 times (50 trials in total), and report the average SR (Avg. SR). Additionally, for PointVLM evaluation, we use accuracy (Acc), defined as the ratio of correctly predicted points within the ground truth mask to the total predicted points.

\textbf{Implementation Details}  
For Reasoning-VLM, we utilize GPT-4o to interpret high-level human instructions and make spatial decisions. The Pointing-VLM employs the NaviAfford model, trained on the 1.0M Spatial Perception Object Affordances dataset, initialized with pre-trained Qwen2.5-VL-7B weights and fully fine-tuned as described in~\cite{zheng2024llamafactory}. Experiments are conducted on four H100 GPUs using AdamW as the optimizer, with a learning rate of \(10^{-5}\) for one epoch. Each GPU processes a batch size of 4, with gradient accumulation set to 2 steps, resulting in an effective batch size of 32. To validate cross-embodiment capability, we deploy the system on both the RealMan wheeled robot and the Unitree Go2 quadruped robot, each equipped with Intel RealSense D435i cameras for RGB-D perception.

\textbf{Baseline Models}  
Existing navigation methods often struggle with long-horizon tasks involving high-level human instructions. To ensure fair comparison, we adapt their task formulations by modifying the instruction format for explicit guidance: “You need to complete the following instructions: I want to drink coffee. Find the target object to complete the instructions and stop near it.” Additionally, we provide baseline models with top-down global 3D scene information.
We evaluate three types of baseline models: (1) closed-source general VLMs, including GPT-4o~\cite{hurst2024gpt}, Claude-3.5-Sonnet~\cite{anthropic2024claude}, and Qwen-VL-Max~\cite{bai2025qwen2}; (2) open-source general-purpose VLMs, such as Janus-Pro-7B~\cite{chen2025janus}, Qwen2.5-VL-7B~\cite{bai2025qwen2}, and LLaVA-Next-7B~\cite{liu2024llavanext}; and (3) navigation-specific methods, including NaVid~\cite{zhang2024navid}, NaVILA~\cite{cheng2024navila}, and MapNav~\cite{zhang2025mapnav}, which require adaptation for our long-horizon navigation tasks.

\begin{table}[t]
\renewcommand{\arraystretch}{1.3}
\centering
\caption{\textbf{Ablation Study on Annotation Components.}}
\label{tab:ablation_annotation}
\resizebox{0.47\textwidth}{!}{%
\begin{tabular}{l|cc|cc|c}
\hline
\multirow{2}{*}{\textbf{Annotation Variants}} & \multicolumn{2}{c|}{\textbf{Tea Room}} & \multicolumn{2}{c|}{\textbf{Workstation}} & \multirow{2}{*}{\textbf{Avg. SR$\uparrow$}} \\
\cline{2-5}
& \textbf{NE$\downarrow$} & \textbf{SR$\uparrow$} & \textbf{NE$\downarrow$} & \textbf{SR$\uparrow$} & \\
\hline
\NickName w/o Map & 4.13 & 32.0\% & 5.88 & 40.0\% & 36.0\% \\
\NickName w/o Annotation & 3.32 & 36.0\% & 2.78 & 44.0\% & 40.0\% \\
\NickName w/o Room-level Annotation & 3.21 & 36.0\% & 3.01 & 40.0\% & 38.0\% \\
\hline
\rowcolor{yellow!8}\textbf{\NickName w/ Full Annotation (Ours)} & \textbf{1.89} & \textbf{60.0\%} & \textbf{1.56} & \textbf{76.0\%} & \textbf{68.0\%} \\
\hline
\end{tabular}%
}
\vspace{-1em}
\end{table}

\subsection{Comparisons with SOTA Methods}  
As shown in Tab.~\ref{tab:navigation_results}, \textit{\textbf{\NickName}} significantly outperforms existing state-of-the-art methods across all evaluation scenarios, achieving a 41.2 percentage point increase in success rate (SR) with an average of 66.4\%, compared to the best baseline, MapNav~\cite{zhang2025mapnav}, at 25.2\%. Specifically, \textit{\textbf{\NickName}} improves SR by 46.0\% in Conference Room A (72.0\% \textit{vs.} 26.0\%), 40.0\% in Conference Room B (64.0\% \textit{vs.} 24.0\%), 34.0\% in Tea Room (60.0\% \textit{vs.} 26.0\%), 48.0\% in Workstation (76.0\% \textit{vs.} 28.0\%), and 38.0\% in Balcony (60.0\% \textit{vs.} 22.0\%). It also significantly reduces navigation error (NE) in all scenarios: by 5.98m (1.23m \textit{vs.} 7.21m) in Conference Room A, 6.49m (1.45m \textit{vs.} 7.94m) in Conference Room B, 7.23m (1.89m \textit{vs.} 9.12m) in Tea Room, 5.22m (1.56m \textit{vs.} 6.78m) in Workstation, and 6.11m (1.34m \textit{vs.} 7.45m) in Balcony. While general-purpose VLMs (both closed-source and open-source) often achieve near-zero success rates in this challenging long-horizon navigation task, our hierarchical approach effectively bridges the gap between high-level command understanding and accurate spatial navigation in real-world environments.

\vspace{-3pt}
\subsection{Ablation Study}

\begin{table}[t]
\renewcommand{\arraystretch}{1.3}
\centering
\caption{\textbf{Ablation Study on Different Reasoning-VLMs.}}
\label{tab:ablation_reasonvlm}
\resizebox{0.46\textwidth}{!}{%
\begin{tabular}{l|cc|cc|c}
\hline
\multirow{2}{*}{\textbf{Reasoning-VLMs}} & \multicolumn{2}{c|}{\textbf{Tea Room}} & \multicolumn{2}{c|}{\textbf{Workstation}} & \multirow{2}{*}{\textbf{Avg. SR$\uparrow$}} \\
\cline{2-5}
& \textbf{NE$\downarrow$} & \textbf{SR$\uparrow$} & \textbf{NE$\downarrow$} & \textbf{SR$\uparrow$} & \\
\hline
\multicolumn{6}{c}{\textit{Open-source}} \\
\hline
\NickName w/ Qwen2.5-VL-72B~\cite{bai2025qwen2} & 2.67 & 52.0\% & 2.23 & 66.0\% & 59.0\% \\
\NickName w/ Qwen2.5-VL-7B~\cite{bai2025qwen2} & 3.12 & 38.0\% & 2.89 & 42.0\% & 40.0\% \\
\NickName w/ Janus-Pro-7B~\cite{chen2025janus} & 3.78 & 30.0\% & 3.56 & 34.0\% & 32.0\% \\
\NickName w/ LLaVA-NeXT-7B~\cite{liu2024llavanext} & 3.45 & 34.0\% & 3.23 & 38.0\% & 36.0\% \\

\hline
\multicolumn{6}{c}{\textit{Closed-source}} \\
\hline
\NickName w/ Claude-3.5-Sonnet~\cite{anthropic2024claude} & 2.01 & 58.0\% & 1.68 & 72.0\% & 65.0\% \\
\NickName w/ Qwen-VL-Max~\cite{bai2025qwen2} & 2.34 & 54.0\% & 2.12 & 68.0\% & 61.0\% \\
\rowcolor{green!10}\textbf{\NickName w/ GPT-4o~\cite{hurst2024gpt} (Ours)} & \textbf{1.89} & \textbf{60.0\%} & \textbf{1.56} & \textbf{76.0\%} & \textbf{68.0\%} \\

\hline
\end{tabular}%
}
\vspace{-1.5em}
\end{table}


\begin{figure*}[!t]
\centering
\includegraphics[width=0.96\textwidth]{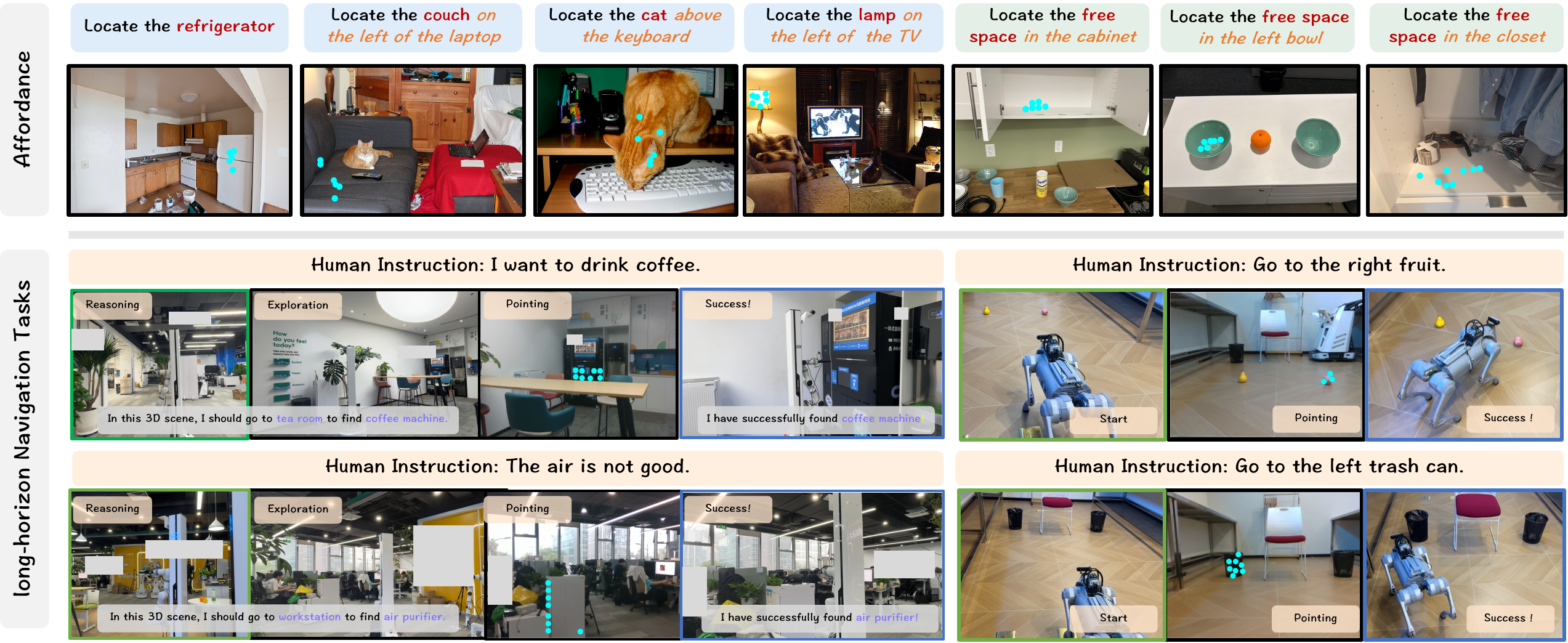} 
\vspace{-3pt}
\caption{\textbf{Qualitative analysis on \textit{NaviAfford} and \textit{\NickName}.} Affordance visualization includes the performance of \textit{\textbf{NaviAfford}} model on object affordance and spatial affordance. Long-horizon navigation tasks visualization includes the performance of \textit{\textbf{\NickName}} hierarchical system in real-world environments and its cross-embodiment deployment capabilities.}
\label{fig5}
\vspace{-1em}
\end{figure*}

\textbf{Effect of Annotation}  
To evaluate our annotation strategy, we conducted ablation studies in the tea room and workstation. Results in Tab.~\ref{tab:ablation_annotation} demonstrate the importance of semantic annotation in long-horizon navigation. Compared to the \textit{\textbf{\NickName}} w/o map, \textit{\textbf{\NickName}} w/ full annotation (ours) shows a 28.0\% improvement in tea room (60.0\% \textit{vs.} 32.0\%) and a 36.0\% improvement in the workstation (76.0\% \textit{vs.} 40.0\%), resulting in an average SR improvement of 32.0\% (68.0\% \textit{vs.} 36.0\%). When compared to \textit{\textbf{\NickName}} w/o annotation, we see a 24.0\% improvement in the tea room (60.0\% \textit{vs.} 36.0\%) and a 32.0\% improvement in the workstation (76.0\% \textit{vs.} 44.0\%), leading to an average enhancement of 28.0\%. Against \textit{\textbf{\NickName}} w/o room-level annotation, our strategy yields a 24.0\% improvement in the tea room (60.0\% \textit{vs.} 36.0\%) and a 36.0\% improvement in the workstation (76.0\% \textit{vs.} 40.0\%), with an average enhancement of 30.0\%. These findings confirm that detailed semantic annotations improve Reasoning-VLMs' understanding of spatial relationships.

\textbf{Effect of Reasoning-VLMs}  
To evaluate the impact of different Reasoning-VLM models on navigation performance, we conducted ablation studies in the tea room and workstation scenarios. Results in Tab.~\ref{tab:ablation_reasonvlm} reveal significant differences among VLM architectures. Our GPT-4o~\cite{hurst2024gpt}-based Reasoning-VLM achieves the highest success rate (SR) of 68.0\%. Closed-source models like Claude-3.5-Sonnet~\cite{anthropic2024claude} and Qwen-VL-Max~\cite{bai2025qwen2} show decreases of 3.0\% (65.0\%) and 7.0\% (61.0\%), respectively. Open-source models, such as Qwen2.5-VL-72B~\cite{bai2025qwen2}, drop by 9.0\% (59.0\%). Smaller 7B models, including Qwen2.5-VL-7B~\cite{bai2025qwen2}, Janus-Pro-7B~\cite{chen2025janus}, and LLaVA-NeXT-7B~\cite{liu2024llavanext}, exhibit declines of 28.0\%, 32.0\%, and 36.0\%, respectively. These findings underscore the importance of reasoning capabilities in complex spatial tasks.

\begin{table}[t]
\renewcommand{\arraystretch}{1.3}
\centering
\caption{\textbf{Ablation Study on Different Pointing-VLMs.}}
\label{tab:ablation_pointvlm}
\resizebox{0.48\textwidth}{!}{%
\begin{tabular}{l|c|cc|c|c}
\hline
\textbf{Pointing-VLMs} & \textbf{Params} & \textbf{Obj. Aff.$\uparrow$} & \textbf{Spa. Aff.$\uparrow$} & \textbf{Avg. Acc$\uparrow$} & \textbf{Nav. SR$\uparrow$} \\
\hline
\multicolumn{6}{c}{\textit{Closed-source}} \\
\hline
\NickName w/ GPT-4o~\cite{hurst2024gpt} & - & 21.3\% & 25.1\% & 23.2\% & 32.0\% \\
\NickName w/ Claude-3.5-Sonnet~\cite{anthropic2024claude} & - & 20.1\% & 22.8\% & 21.5\% & 28.0\% \\
\NickName w/ Qwen-VL-Max~\cite{bai2025qwen2} & - & 18.8\% & 21.5\% & 20.2\% & 20.0\% \\
\hline
\multicolumn{6}{c}{\textit{Open-source}} \\
\hline
\NickName w/ Qwen2.5-VL~\cite{bai2025qwen2} & 72B & 15.2\% & 18.7\% & 17.0\% & 16.0\% \\
\NickName w/ Qwen2.5-VL~\cite{bai2025qwen2} & 7B & 6.2\% & 15.3\% & 10.8\% & 12.0\% \\
\NickName w/ Janus-Pro~\cite{chen2025janus} & 7B & 2.8\% & 3.1\% & 2.95\% & 6.0\% \\
\NickName w/ LLaVA-NeXT~\cite{liu2024llavanext} & 7B & 3.1\% & 0.8\% & 2.0\% & 4.0\% \\
\hline
\multicolumn{6}{c}{\textit{Specific}} \\
\hline
\NickName w/ RoboPoint~\cite{yuan2024robopoint} & 13B & 55.9\% & 44.5\% & 50.2\% & 57.5\% \\
\rowcolor{red!8}\textbf{\NickName w/ NaviAfford (Ours)} & \textbf{7B} & \textbf{70.8\%} & \textbf{55.6\%} & \textbf{63.2\%} & \textbf{68.0\%} \\
\hline
\end{tabular}%
}
\vspace{-1.5em}
\end{table}

\textbf{Effect of Pointing-VLMs}  
To evaluate the effectiveness of different Pointing-VLMs for object localization, we compare the NaviAfford model with baseline methods. Results in Tab.~\ref{tab:ablation_pointvlm} highlight NaviAfford's superior performance on the affordance understanding benchmark, achieving a 13.0\% improvement in average affordance accuracy over the previous state-of-the-art RoboPoint~\cite{yuan2024robopoint} (70.8\% \textit{vs.} 55.9\%). This strong affordance understanding translates to enhanced navigation performance, with NaviAfford showing a 10.5\% increase in success rate (SR) over RoboPoint (68.0\% \textit{vs.} 57.5\%), a 36.0\% improvement over GPT-4o~\cite{hurst2024gpt} (68.0\% \textit{vs.} 32.0\%), and a 52.0\% boost over the best open-source model, Qwen2.5-VL-72B~\cite{bai2025qwen2} (68.0\% \textit{vs.} 16.0\%). These results demonstrate that our spatial affordance training effectively bridges accurate object localization and practical navigation execution.

\subsection{Qualitative Analysis}  
We perform a qualitative evaluation to illustrate the capabilities of \textit{\textbf{\NickName}} in affordance understanding, navigation, and cross-embodiment deployment, as shown in Fig.~\ref{fig5}. Affordance visualizations highlight NaviAfford’s spatial awareness, accurately identifying references like “the sofa to the left of the laptop” and “the empty space in the closet,” while localizing objects in cluttered settings. Long-horizon navigation visualizations demonstrate our framework’s systematic approach, tracing clear reasoning from instruction parsing (\textit{e.g.}, “I want coffee”) to goal achievement in multi-room environments. Cross-embodiment experiments showcase the versatility of \textit{\textbf{\NickName}}, achieving consistent performance on quadruped robots in tasks such as “walk to the fruit on the right.” These findings confirm the adaptability of our approach across various robotic platforms.

\vspace{-3pt}
\section{Conclusion}  
This paper presents \textit{\textbf{\NickName}}, a hierarchical framework that bridges embodied navigation research with real-world human needs, enabling robots to interpret high-level instructions and navigate complex environments. Our approach divides navigation into two stages: a global policy utilizing Reasoning-VLM for instruction parsing and target identification, and a local strategy with our \textit{\textbf{NaviAfford}} model for precise object localization. Extensive experiments show that \textit{\textbf{\NickName}} outperforms state-of-the-art methods in managing complex spatial relations and open-vocabulary pointing. Successful deployment on wheeled and quadruped robots highlights its versatility. Future work will focus on enhancing adaptability in dynamic environments and integrating additional sensory inputs to boost navigation performance.

\bibliography{aaai2026}

\begin{thebibliography}{57}
\providecommand{\natexlab}[1]{#1}

\bibitem[{Anthropic(2024)}]{anthropic2024claude}
Anthropic. 2024.
\newblock The claude 3 model family: Opus, sonnet, haiku.
\newblock \emph{https://docs.anthropic.com/zh-CN/release-notes/claude-apps}.

\bibitem[{Bai et~al.(2025)Bai, Chen, Liu, Wang, Ge, Song, Dang, Wang, Wang, Tang et~al.}]{bai2025qwen2}
Bai, S.; Chen, K.; Liu, X.; Wang, J.; Ge, W.; Song, S.; Dang, K.; Wang, P.; Wang, S.; Tang, J.; et~al. 2025.
\newblock Qwen2. 5-vl technical report.
\newblock \emph{arXiv preprint arXiv:2502.13923}.

\bibitem[{Beyer et~al.(2024)Beyer, Steiner, Pinto, Kolesnikov, Wang, Salz, Neumann, Alabdulmohsin, Tschannen, Bugliarello et~al.}]{beyer2024paligemma}
Beyer, L.; Steiner, A.; Pinto, A.~S.; Kolesnikov, A.; Wang, X.; Salz, D.; Neumann, M.; Alabdulmohsin, I.; Tschannen, M.; Bugliarello, E.; et~al. 2024.
\newblock Paligemma: A versatile 3b vlm for transfer.
\newblock \emph{arXiv preprint arXiv:2407.07726}.

\bibitem[{Cai et~al.(2024{\natexlab{a}})Cai, Huang, Cheng, Long, Gao, Sun, and Dong}]{cai2024bridging}
Cai, W.; Huang, S.; Cheng, G.; Long, Y.; Gao, P.; Sun, C.; and Dong, H. 2024{\natexlab{a}}.
\newblock Bridging zero-shot object navigation and foundation models through pixel-guided navigation skill.
\newblock In \emph{IEEE International Conference on Robotics and Automation}, 5228--5234. IEEE.

\bibitem[{Cai et~al.(2024{\natexlab{b}})Cai, Ponomarenko, Yuan, Li, Yang, Dong, and Zhao}]{cai2024spatialbot}
Cai, W.; Ponomarenko, I.; Yuan, J.; Li, X.; Yang, W.; Dong, H.; and Zhao, B. 2024{\natexlab{b}}.
\newblock Spatialbot: Precise spatial understanding with vision language models.
\newblock \emph{arXiv preprint arXiv:2406.13642}.

\bibitem[{Chattopadhyay et~al.(2021)Chattopadhyay, Hoffman, Mottaghi, and Kembhavi}]{chattopadhyay2021robustnav}
Chattopadhyay, P.; Hoffman, J.; Mottaghi, R.; and Kembhavi, A. 2021.
\newblock Robustnav: Towards benchmarking robustness in embodied navigation.
\newblock In \emph{Proceedings of the IEEE/CVF International Conference on Computer Vision}, 15691--15700.

\bibitem[{Chen et~al.(2024{\natexlab{a}})Chen, Xu, Kirmani, Ichter, Sadigh, Guibas, and Xia}]{chen2024spatialvlm}
Chen, B.; Xu, Z.; Kirmani, S.; Ichter, B.; Sadigh, D.; Guibas, L.; and Xia, F. 2024{\natexlab{a}}.
\newblock Spatialvlm: Endowing vision-language models with spatial reasoning capabilities.
\newblock In \emph{Proceedings of the IEEE/CVF Conference on Computer Vision and Pattern Recognition}, 14455--14465.

\bibitem[{Chen et~al.(2024{\natexlab{b}})Chen, Lin, Xu, Chai, Liang, and Wong}]{chen2024mapgpt}
Chen, J.; Lin, B.; Xu, R.; Chai, Z.; Liang, X.; and Wong, K.-Y. 2024{\natexlab{b}}.
\newblock MapGPT: Map-Guided Prompting with Adaptive Path Planning for Vision-and-Language Navigation.
\newblock In \emph{Proceedings of the 62nd Annual Meeting of the Association for Computational Linguistics}, 9796--9810.

\bibitem[{Chen et~al.(2025)Chen, Wu, Liu, Pan, Liu, Xie, Yu, and Ruan}]{chen2025janus}
Chen, X.; Wu, Z.; Liu, X.; Pan, Z.; Liu, W.; Xie, Z.; Yu, X.; and Ruan, C. 2025.
\newblock Janus-pro: Unified multimodal understanding and generation with data and model scaling.
\newblock \emph{arXiv preprint arXiv:2501.17811}.

\bibitem[{Cheng et~al.(2025)Cheng, Ji, Yang, Gongye, Zou, Kautz, B{\i}y{\i}k, Yin, Liu, and Wang}]{cheng2024navila}
Cheng, A.-C.; Ji, Y.; Yang, Z.; Gongye, Z.; Zou, X.; Kautz, J.; B{\i}y{\i}k, E.; Yin, H.; Liu, S.; and Wang, X. 2025.
\newblock Navila: Legged robot vision-language-action model for navigation.
\newblock In \emph{Robotics: Science and Systems}.

\bibitem[{Cheng et~al.(2024)Cheng, Yin, Fu, Guo, Yang, Kautz, Wang, and Liu}]{cheng2024spatialrgpt}
Cheng, A.-C.; Yin, H.; Fu, Y.; Guo, Q.; Yang, R.; Kautz, J.; Wang, X.; and Liu, S. 2024.
\newblock Spatialrgpt: Grounded spatial reasoning in vision-language models.
\newblock \emph{Advances in Neural Information Processing Systems}, 37: 135062--135093.

\bibitem[{Doveh et~al.(2024)Doveh, Perek, Mirza, Lin, Alfassy, Arbelle, Ullman, and Karlinsky}]{doveh2024towards}
Doveh, S.; Perek, S.; Mirza, M.~J.; Lin, W.; Alfassy, A.; Arbelle, A.; Ullman, S.; and Karlinsky, L. 2024.
\newblock Towards Multimodal In-context Learning for Vision and Language Models.
\newblock In \emph{European Conference on Computer Vision}, 250--267. Springer.

\bibitem[{Gao et~al.(2025)Gao, Jin, Peng, Zhang, Deng, Li, Wang, and Liu}]{gao2025octonav}
Gao, C.; Jin, L.; Peng, X.; Zhang, J.; Deng, Y.; Li, A.; Wang, H.; and Liu, S. 2025.
\newblock OctoNav: Towards Generalist Embodied Navigation.
\newblock \emph{arXiv preprint arXiv:2506.09839}.

\bibitem[{Gong et~al.(2025)Gong, Li, Hu, Qiu, Kong, Zhang, Ding, Zhang, and Liang}]{gong2025stairway}
Gong, Z.; Li, R.; Hu, T.; Qiu, R.; Kong, L.; Zhang, L.; Ding, Y.; Zhang, L.; and Liang, J. 2025.
\newblock Stairway to Success: Zero-Shot Floor-Aware Object-Goal Navigation via LLM-Driven Coarse-to-Fine Exploration.
\newblock \emph{arXiv preprint arXiv:2505.23019}.

\bibitem[{Hao et~al.(2025{\natexlab{a}})Hao, Zhang, Li, Cao, Hao, Cui, and Wang}]{hao2025tla}
Hao, P.; Zhang, C.; Li, D.; Cao, X.; Hao, X.; Cui, S.; and Wang, S. 2025{\natexlab{a}}.
\newblock Tla: Tactile-language-action model for contact-rich manipulation.
\newblock \emph{arXiv preprint arXiv:2503.08548}.

\bibitem[{Hao et~al.(2025{\natexlab{b}})Hao, Diao, Wei, Yang, Hao, Yin, Zhang, Li, Zhao, and Liu}]{hao2025mapfusion}
Hao, X.; Diao, Y.; Wei, M.; Yang, Y.; Hao, P.; Yin, R.; Zhang, H.; Li, W.; Zhao, S.; and Liu, Y. 2025{\natexlab{b}}.
\newblock MapFusion: A novel BEV feature fusion network for multi-modal map construction.
\newblock \emph{Information Fusion}, 119: 103018.

\bibitem[{Hao et~al.(2025{\natexlab{c}})Hao, Kong, Yin, Wang, Zhang, Diao, and Zhao}]{hao2025safemap}
Hao, X.; Kong, L.; Yin, R.; Wang, P.; Zhang, J.; Diao, Y.; and Zhao, S. 2025{\natexlab{c}}.
\newblock SafeMap: Robust HD Map Construction from Incomplete Observations.
\newblock In \emph{Forty-second International Conference on Machine Learning}.

\bibitem[{Hao et~al.(2024{\natexlab{a}})Hao, Li, Zhang, Li, Yin, Jung, Park, Yoo, Zhao, and Zhang}]{hao2024mapdistill}
Hao, X.; Li, R.; Zhang, H.; Li, D.; Yin, R.; Jung, S.; Park, S.-I.; Yoo, B.; Zhao, H.; and Zhang, J. 2024{\natexlab{a}}.
\newblock Mapdistill: Boosting efficient camera-based hd map construction via camera-lidar fusion model distillation.
\newblock In \emph{European Conference on Computer Vision}, 166--183. Springer.

\bibitem[{Hao et~al.(2025{\natexlab{d}})Hao, Liu, Zhao, Ji, Wei, Zhao, Kong, Yin, and Liu}]{hao2025msc}
Hao, X.; Liu, G.; Zhao, Y.; Ji, Y.; Wei, M.; Zhao, H.; Kong, L.; Yin, R.; and Liu, Y. 2025{\natexlab{d}}.
\newblock MSC-Bench: Benchmarking and Analyzing Multi-Sensor Corruption for Driving Perception.
\newblock \emph{arXiv preprint arXiv:2501.01037}.

\bibitem[{Hao et~al.(2024{\natexlab{b}})Hao, Zhang, Yang, Zhou, Jung, Park, and Yoo}]{hao2024mbfusion}
Hao, X.; Zhang, H.; Yang, Y.; Zhou, Y.; Jung, S.; Park, S.-I.; and Yoo, B. 2024{\natexlab{b}}.
\newblock Mbfusion: A new multi-modal bev feature fusion method for hd map construction.
\newblock In \emph{IEEE International Conference on Robotics and Automation}, 15922--15928. IEEE.

\bibitem[{Hong et~al.(2021)Hong, Wu, Qi, Rodriguez-Opazo, and Gould}]{hong2021vln}
Hong, Y.; Wu, Q.; Qi, Y.; Rodriguez-Opazo, C.; and Gould, S. 2021.
\newblock Vln bert: A recurrent vision-and-language bert for navigation.
\newblock In \emph{Proceedings of the IEEE/CVF conference on Computer Vision and Pattern Recognition}, 1643--1653.

\bibitem[{Hurst et~al.(2024)Hurst, Lerer, Goucher, Perelman, Ramesh, Clark, Ostrow, Welihinda, Hayes, Radford et~al.}]{hurst2024gpt}
Hurst, A.; Lerer, A.; Goucher, A.~P.; Perelman, A.; Ramesh, A.; Clark, A.; Ostrow, A.; Welihinda, A.; Hayes, A.; Radford, A.; et~al. 2024.
\newblock Gpt-4o system card.
\newblock \emph{arXiv preprint arXiv:2410.21276}.

\bibitem[{Ji et~al.(2025)Ji, Tan, Shi, Hao, Zhang, Zhang, Wang, Zhao, Mu, An et~al.}]{ji2025robobrain}
Ji, Y.; Tan, H.; Shi, J.; Hao, X.; Zhang, Y.; Zhang, H.; Wang, P.; Zhao, M.; Mu, Y.; An, P.; et~al. 2025.
\newblock Robobrain: A unified brain model for robotic manipulation from abstract to concrete.
\newblock In \emph{Proceedings of the Computer Vision and Pattern Recognition Conference}, 1724--1734.

\bibitem[{Li et~al.(2024{\natexlab{a}})Li, Jin, Sun, Yu, Shi, Hao, Hao, Liu, Sun, Zhang et~al.}]{li2024foundation}
Li, D.; Jin, Y.; Sun, Y.; Yu, H.; Shi, J.; Hao, X.; Hao, P.; Liu, H.; Sun, F.; Zhang, J.; et~al. 2024{\natexlab{a}}.
\newblock What foundation models can bring for robot learning in manipulation: A survey.
\newblock \emph{arXiv preprint arXiv:2404.18201}.

\bibitem[{Li et~al.(2024{\natexlab{b}})Li, Zhang, Zhang, Zhang, Li, Li, Ma, and Li}]{liu2024llavanext}
Li, F.; Zhang, R.; Zhang, H.; Zhang, Y.; Li, B.; Li, W.; Ma, Z.; and Li, C. 2024{\natexlab{b}}.
\newblock Llava-next-interleave: Tackling multi-image, video, and 3d in large multimodal models.
\newblock \emph{arXiv preprint arXiv:2407.07895}.

\bibitem[{Lin et~al.(2025)Lin, Nie, Wei, Chen, Ma, Han, Xu, Chang, and Liang}]{lin2025navcot}
Lin, B.; Nie, Y.; Wei, Z.; Chen, J.; Ma, S.; Han, J.; Xu, H.; Chang, X.; and Liang, X. 2025.
\newblock Navcot: Boosting llm-based vision-and-language navigation via learning disentangled reasoning.
\newblock \emph{IEEE Transactions on Pattern Analysis and Machine Intelligence}.

\bibitem[{Liu et~al.(2023)Liu, Wang, Ye, Chong, Zhou, and Hua}]{liu2023qilin}
Liu, J.; Wang, Z.; Ye, Q.; Chong, D.; Zhou, P.; and Hua, Y. 2023.
\newblock Qilin-med-vl: Towards chinese large vision-language model for general healthcare.
\newblock \emph{arXiv preprint arXiv:2310.17956}.

\bibitem[{Liu et~al.(2024)Liu, Zhang, Gao, Wang, and Wang}]{liu2024vision}
Liu, S.; Zhang, J.; Gao, R.~X.; Wang, X.~V.; and Wang, L. 2024.
\newblock Vision-language model-driven scene understanding and robotic object manipulation.
\newblock In \emph{IEEE 20th International Conference on Automation Science and Engineering}, 21--26. IEEE.

\bibitem[{Liu et~al.(2025)Liu, Chi, Wu, Zhang, Hu, Zhang, Zhang, Wu, Cao, Huang et~al.}]{liu2025spatialcot}
Liu, Y.; Chi, D.; Wu, S.; Zhang, Z.; Hu, Y.; Zhang, L.; Zhang, Y.; Wu, S.; Cao, T.; Huang, G.; et~al. 2025.
\newblock SpatialCoT: Advancing Spatial Reasoning through Coordinate Alignment and Chain-of-Thought for Embodied Task Planning.
\newblock \emph{arXiv preprint arXiv:2501.10074}.

\bibitem[{Long et~al.(2025)Long, Cai, Wang, Zhan, and Dong}]{long2024instructnav}
Long, Y.; Cai, W.; Wang, H.; Zhan, G.; and Dong, H. 2025.
\newblock InstructNav: Zero-shot System for Generic Instruction Navigation in Unexplored Environment.
\newblock In \emph{Conference on Robot Learning}, 2049--2060.

\bibitem[{Long et~al.(2024)Long, Li, Cai, and Dong}]{long2024discuss}
Long, Y.; Li, X.; Cai, W.; and Dong, H. 2024.
\newblock Discuss before moving: Visual language navigation via multi-expert discussions.
\newblock In \emph{IEEE International Conference on Robotics and Automation}, 17380--17387. IEEE.

\bibitem[{Luo et~al.(2023)Luo, Zhou, Ren, Chen, Sun, and Ji}]{luo2023cheap}
Luo, G.; Zhou, Y.; Ren, T.; Chen, S.; Sun, X.; and Ji, R. 2023.
\newblock Cheap and quick: Efficient vision-language instruction tuning for large language models.
\newblock \emph{Advances in Neural Information Processing Systems}, 36: 29615--29627.

\bibitem[{Morad et~al.(2021)Morad, Mecca, Poudel, Liwicki, and Cipolla}]{morad2021embodied}
Morad, S.~D.; Mecca, R.; Poudel, R.~P.; Liwicki, S.; and Cipolla, R. 2021.
\newblock Embodied visual navigation with automatic curriculum learning in real environments.
\newblock \emph{IEEE Robotics and Automation Letters}, 6(2): 683--690.

\bibitem[{O’Neill et~al.(2024)O’Neill, Rehman, Maddukuri, Gupta, Padalkar, Lee, Pooley, Gupta, Mandlekar, Jain et~al.}]{o2024open}
O’Neill, A.; Rehman, A.; Maddukuri, A.; Gupta, A.; Padalkar, A.; Lee, A.; Pooley, A.; Gupta, A.; Mandlekar, A.; Jain, A.; et~al. 2024.
\newblock Open x-embodiment: Robotic learning datasets and rt-x models: Open x-embodiment collaboration 0.
\newblock In \emph{IEEE International Conference on Robotics and Automation}, 6892--6903. IEEE.

\bibitem[{Qi et~al.(2025)Qi, Zhang, Yu, Wang, and Zhao}]{qi2025vln}
Qi, Z.; Zhang, Z.; Yu, Y.; Wang, J.; and Zhao, H. 2025.
\newblock VLN-R1: Vision-Language Navigation via Reinforcement Fine-Tuning.
\newblock \emph{arXiv preprint arXiv:2506.17221}.

\bibitem[{Ramrakhya et~al.(2023)Ramrakhya, Batra, Wijmans, and Das}]{ramrakhya2023pirlnav}
Ramrakhya, R.; Batra, D.; Wijmans, E.; and Das, A. 2023.
\newblock Pirlnav: Pretraining with imitation and rl finetuning for objectnav.
\newblock In \emph{Proceedings of the IEEE/CVF Conference on Computer Vision and Pattern Recognition}, 17896--17906.

\bibitem[{Tan et~al.(2025)Tan, Ji, Hao, Lin, Wang, Wang, and Zhang}]{tan2025reason}
Tan, H.; Ji, Y.; Hao, X.; Lin, M.; Wang, P.; Wang, Z.; and Zhang, S. 2025.
\newblock Reason-rft: Reinforcement fine-tuning for visual reasoning.
\newblock \emph{arXiv preprint arXiv:2503.20752}.

\bibitem[{Tang et~al.(2025)Tang, Zhang, Hao, Wang, Wu, Wang, and Zhang}]{tang2025affordgrasp}
Tang, Y.; Zhang, S.; Hao, X.; Wang, P.; Wu, J.; Wang, Z.; and Zhang, S. 2025.
\newblock Affordgrasp: In-context affordance reasoning for open-vocabulary task-oriented grasping in clutter.
\newblock \emph{arXiv preprint arXiv:2503.00778}.

\bibitem[{Truong, Chernova, and Batra(2021)}]{truong2021bi}
Truong, J.; Chernova, S.; and Batra, D. 2021.
\newblock Bi-directional domain adaptation for sim2real transfer of embodied navigation agents.
\newblock \emph{IEEE Robotics and Automation Letters}, 6(2): 2634--2641.

\bibitem[{Wang et~al.(2023)Wang, Chen, Chen, Wu, Zhu, Zeng, Luo, Lu, Zhou, Qiao et~al.}]{wang2023visionllm}
Wang, W.; Chen, Z.; Chen, X.; Wu, J.; Zhu, X.; Zeng, G.; Luo, P.; Lu, T.; Zhou, J.; Qiao, Y.; et~al. 2023.
\newblock Visionllm: Large language model is also an open-ended decoder for vision-centric tasks.
\newblock \emph{Advances in Neural Information Processing Systems}, 36: 61501--61513.

\bibitem[{Wasserman et~al.(2024)Wasserman, Chowdhary, Gupta, and Jain}]{wasserman2024exploitation}
Wasserman, J.; Chowdhary, G.; Gupta, A.; and Jain, U. 2024.
\newblock Exploitation-guided exploration for semantic embodied navigation.
\newblock In \emph{IEEE International Conference on Robotics and Automation}, 2901--2908. IEEE.

\bibitem[{Wu et~al.(2025)Wu, Guan, Feng, Liu, Wu, Wang, Wu, and Tan}]{wu2025reinforcing}
Wu, J.; Guan, J.; Feng, K.; Liu, Q.; Wu, S.; Wang, L.; Wu, W.; and Tan, T. 2025.
\newblock Reinforcing Spatial Reasoning in Vision-Language Models with Interwoven Thinking and Visual Drawing.
\newblock \emph{arXiv preprint arXiv:2506.09965}.

\bibitem[{Yu, Kasaei, and Cao(2023)}]{yu2023l3mvn}
Yu, B.; Kasaei, H.; and Cao, M. 2023.
\newblock L3mvn: Leveraging large language models for visual target navigation.
\newblock In \emph{IEEE/RSJ International Conference on Intelligent Robots and Systems}, 3554--3560. IEEE.

\bibitem[{Yuan et~al.(2025)Yuan, Duan, Blukis, Pumacay, Krishna, Murali, Mousavian, and Fox}]{yuan2024robopoint}
Yuan, W.; Duan, J.; Blukis, V.; Pumacay, W.; Krishna, R.; Murali, A.; Mousavian, A.; and Fox, D. 2025.
\newblock RoboPoint: A Vision-Language Model for Spatial Affordance Prediction in Robotics.
\newblock In \emph{Conference on Robot Learning}, 4005--4020.

\bibitem[{Zhai et~al.(2024)Zhai, Bai, Lin, Pan, Tong, Zhou, Suhr, Xie, LeCun, Ma et~al.}]{zhai2024fine}
Zhai, S.; Bai, H.; Lin, Z.; Pan, J.; Tong, P.; Zhou, Y.; Suhr, A.; Xie, S.; LeCun, Y.; Ma, Y.; et~al. 2024.
\newblock Fine-tuning large vision-language models as decision-making agents via reinforcement learning.
\newblock \emph{Advances in neural information processing systems}, 37: 110935--110971.

\bibitem[{Zhang et~al.(2025{\natexlab{a}})Zhang, Hao, Cao, Hao, Cui, and Wang}]{zhang2025vtla}
Zhang, C.; Hao, P.; Cao, X.; Hao, X.; Cui, S.; and Wang, S. 2025{\natexlab{a}}.
\newblock VTLA: Vision-Tactile-Language-Action Model with Preference Learning for Insertion Manipulation.
\newblock \emph{arXiv preprint arXiv:2505.09577}.

\bibitem[{Zhang et~al.(2024{\natexlab{a}})Zhang, Wang, Wang, Li, Liu, Wei, Wang, Zhang, and Wang}]{zhang2024uni}
Zhang, J.; Wang, K.; Wang, S.; Li, M.; Liu, H.; Wei, S.; Wang, Z.; Zhang, Z.; and Wang, H. 2024{\natexlab{a}}.
\newblock Uni-NaVid: A Video-based Vision-Language-Action Model for Unifying Embodied Navigation Tasks.
\newblock \emph{arXiv preprint arXiv:2412.06224}.

\bibitem[{Zhang et~al.(2024{\natexlab{b}})Zhang, Wang, Xu, Zhou, Hong, Fang, Wu, Zhang, and Wang}]{zhang2024navid}
Zhang, J.; Wang, K.; Xu, R.; Zhou, G.; Hong, Y.; Fang, X.; Wu, Q.; Zhang, Z.; and Wang, H. 2024{\natexlab{b}}.
\newblock Navid: Video-based vlm plans the next step for vision-and-language navigation.
\newblock In \emph{Robotics: Science and Systems}.

\bibitem[{Zhang et~al.(2025{\natexlab{b}})Zhang, Hao, Xu, Zhang, Zhang, Wang, Zhang, Wang, Zhang, and Xu}]{zhang2025mapnav}
Zhang, L.; Hao, X.; Xu, Q.; Zhang, Q.; Zhang, X.; Wang, P.; Zhang, J.; Wang, Z.; Zhang, S.; and Xu, R. 2025{\natexlab{b}}.
\newblock Mapnav: A novel memory representation via annotated semantic maps for vlm-based vision-and-language navigation.
\newblock In \emph{The 63rd Annual Meeting of the Association for Computational Linguistics}.

\bibitem[{Zhang et~al.(2025{\natexlab{c}})Zhang, Wang, Xiao, Zhang, Zhang, Jiang, and Xu}]{zhang2024multi}
Zhang, L.; Wang, H.; Xiao, E.; Zhang, X.; Zhang, Q.; Jiang, Z.; and Xu, R. 2025{\natexlab{c}}.
\newblock Multi-floor zero-shot object navigation policy.
\newblock In \emph{IEEE International Conference on Robotics and Automation}. IEEE.

\bibitem[{Zhang et~al.(2025{\natexlab{d}})Zhang, Zhang, Cui, Sun, Cao, Guo, Han, Zhao, Wang, Sun et~al.}]{zhang2025humanoidpano}
Zhang, Q.; Zhang, Z.; Cui, W.; Sun, J.; Cao, J.; Guo, Y.; Han, G.; Zhao, W.; Wang, J.; Sun, C.; et~al. 2025{\natexlab{d}}.
\newblock Humanoidpano: Hybrid spherical panoramic-lidar cross-modal perception for humanoid robots.
\newblock \emph{arXiv preprint arXiv:2503.09010}.

\bibitem[{Zhang et~al.(2025{\natexlab{e}})Zhang, Hao, Tang, Zhang, Wang, Wang, Ma, and Zhang}]{zhang2025video}
Zhang, S.; Hao, X.; Tang, Y.; Zhang, L.; Wang, P.; Wang, Z.; Ma, H.; and Zhang, S. 2025{\natexlab{e}}.
\newblock Video-CoT: A Comprehensive Dataset for Spatiotemporal Understanding of Videos Based on Chain-of-Thought.
\newblock \emph{arXiv preprint arXiv:2506.08817}.

\bibitem[{Zhao et~al.(2025)Zhao, Yuan, Xu, Hao, Zhang, Wu, Che, Liu, and Tang}]{zhao2025training}
Zhao, Y.; Yuan, J.; Xu, Z.; Hao, X.; Zhang, X.; Wu, K.; Che, Z.; Liu, C.~H.; and Tang, J. 2025.
\newblock Training-free Generation of Temporally Consistent Rewards from VLMs.
\newblock \emph{arXiv preprint arXiv:2507.04789}.

\bibitem[{Zheng et~al.(2024{\natexlab{a}})Zheng, Huang, Zhao, Zhong, and Wang}]{zheng2024towards}
Zheng, D.; Huang, S.; Zhao, L.; Zhong, Y.; and Wang, L. 2024{\natexlab{a}}.
\newblock Towards learning a generalist model for embodied navigation.
\newblock In \emph{Proceedings of the IEEE/CVF Conference on Computer Vision and Pattern Recognition}, 13624--13634.

\bibitem[{Zheng et~al.(2025)Zheng, He, Luo, Zhang, Wang, Shi, and Bai}]{zheng2025railway}
Zheng, X.; He, Y.; Luo, Y.; Zhang, L.; Wang, J.; Shi, T.; and Bai, Y. 2025.
\newblock Railway side slope hazard detection system based on generative models.
\newblock \emph{IEEE Sensors Journal}.

\bibitem[{Zheng et~al.(2024{\natexlab{b}})Zheng, Zhang, Zhang, Ye, Luo, Feng, and Ma}]{zheng2024llamafactory}
Zheng, Y.; Zhang, R.; Zhang, J.; Ye, Y.; Luo, Z.; Feng, Z.; and Ma, Y. 2024{\natexlab{b}}.
\newblock Llamafactory: Unified efficient fine-tuning of 100+ language models.
\newblock \emph{arXiv preprint arXiv:2403.13372}.

\bibitem[{Zhou, Hong, and Wu(2024)}]{zhou2024navgpt}
Zhou, G.; Hong, Y.; and Wu, Q. 2024.
\newblock Navgpt: Explicit reasoning in vision-and-language navigation with large language models.
\newblock In \emph{Proceedings of the AAAI Conference on Artificial Intelligence}, volume~38, 7641--7649.

\end{thebibliography}



\end{document}